\pdfoutput=1

\documentclass[11pt]{article}

\usepackage{acl2024}
\usepackage{times}
\usepackage{latexsym}

\usepackage[T1]{fontenc}

\usepackage[utf8]{inputenc}

\usepackage{microtype}

\usepackage{inconsolata}
\usepackage{microtype}
\usepackage{multirow}
\usepackage{multicol}
\usepackage{makecell}
\usepackage[skins]{tcolorbox}
\usepackage{amsmath}
\usepackage{amsfonts}
\usepackage{adjustbox}
\usepackage{bm}
\usepackage{enumitem}
\usepackage{booktabs}
\usepackage{soul}
\usepackage{bbding}
\usepackage{pifont}

\usepackage[percent]{overpic}

\newenvironment{sanseriffont}{\fontsize{7}{7}\fontfamily{bch}\selectfont}{\par}

\definecolor{pastelblue}{HTML}{A1C9F4}
\definecolor{pastelorange}{HTML}{FFB482}
\definecolor{pastelgreen}{HTML}{8DE5A1}
\definecolor{pastelred}{HTML}{FF9F9B}
\definecolor{pastelpurple}{HTML}{D0BBFF}
\definecolor{pastelgold}{HTML}{FFD700}
\definecolor{pastelpink}{HTML}{FF69B4}

\colorlet{pastelbluemuted}{pastelblue!75}
\colorlet{pastelorangemuted}{pastelorange!75}
\colorlet{pastelgreenmuted}{pastelgreen!75}
\colorlet{pastelredmuted}{pastelred!75}
\colorlet{pastelpurplemuted}{pastelpurple!75}
\colorlet{pastelgoldmuted}{pastelgold!75}
\colorlet{pastelpinkmuted}{pastelpink!75}

\DeclareRobustCommand{\hlomma}[1]{{\sethlcolor{pastelbluemuted}\hl{#1}}}
\DeclareRobustCommand{\hlommb}[1]{{\sethlcolor{pastelorangemuted}\hl{#1}}}
\DeclareRobustCommand{\hlommc}[1]{{\sethlcolor{pastelgreenmuted}\hl{#1}}}
\DeclareRobustCommand{\hlommd}[1]{{\sethlcolor{pastelpurplemuted}\hl{#1}}}
\DeclareRobustCommand{\hlomme}[1]{{\sethlcolor{pastelredmuted}\hl{#1}}}
\DeclareRobustCommand{\hlommf}[1]{{\sethlcolor{pastelgoldmuted}\hl{#1}}}

\def \margin {4pt}
\def \fmargin {-10pt}

\definecolor{red}{RGB}{255, 0, 0}
\definecolor{green}{RGB}{0, 176, 80}

\title{\textsc{ICon}: Improving Inter-Report Consistency in Radiology Report Generation via Lesion-aware Mixup Augmentation}

\author{Wenjun Hou$^{1,2}$, Yi Cheng$^{1\ast}$, Kaishuai Xu$^{1\ast}$, Yan Hu$^{2}$, Wenjie Li$^{1\dagger}$, Jiang Liu$^{2\dagger}$ \\
$^1$Department of Computing, The Hong Kong Polytechnic University, HKSAR, China \\
$^2$Research Institute of Trustworthy Autonomous Systems and \\Department of Computer Science and Engineering, \\
Southern University of Science and Technology, Shenzhen, China \\
\texttt{houwenjun060@gmail.com, \{alyssa.cheng,kaishuaii.xu\}@connect.polyu.hk} \\
\texttt{huy3@sustech.edu.cn, cswjli@comp.polyu.edu.hk, liuj@sustech.edu.cn}
}

\begin{document}
\maketitle
\begingroup\def\thefootnote{$\ast$}\footnotetext{Equal Contribution.}\endgroup
\begingroup\def\thefootnote{$\dagger$}\footnotetext{Corresponding authors.}\endgroup
\begin{abstract}
Previous research on radiology report generation has made significant progress in terms of increasing the clinical accuracy of generated reports. In this paper, we emphasize another crucial quality that it should possess, i.e., \textit{inter-report consistency}, which refers to the capability of generating consistent reports for semantically equivalent radiographs. This quality is even of greater significance than the overall report accuracy in terms of ensuring the system's credibility, as a system prone to providing conflicting results would severely erode users' trust. Regrettably, existing approaches struggle to maintain inter-report consistency, exhibiting biases towards common patterns and susceptibility to lesion variants. To address this issue, we propose \textsc{ICon}, which \underline{I}mproves the inter-report \underline{\textsc{Con}}sistency of radiology report generation. Aiming to enhance the system's ability to capture similarities in semantically equivalent lesions, our approach first involves extracting lesions from input images and examining their characteristics. Then, we introduce a lesion-aware mixup technique to ensure that the representations of the semantically equivalent lesions align with the same attributes, achieved through a linear combination during the training phase. Extensive experiments on three publicly available chest X-ray datasets verify the effectiveness of our approach, both in terms of improving the consistency and accuracy of the generated reports\footnote {Our codes and model checkpoints are available at \url{https://github.com/wjhou/ICon}}.
\end{abstract}
\begin{figure}[t]
	\centering
    \setlength\belowcaptionskip{\fmargin}
    \includegraphics[width=1.0\linewidth]{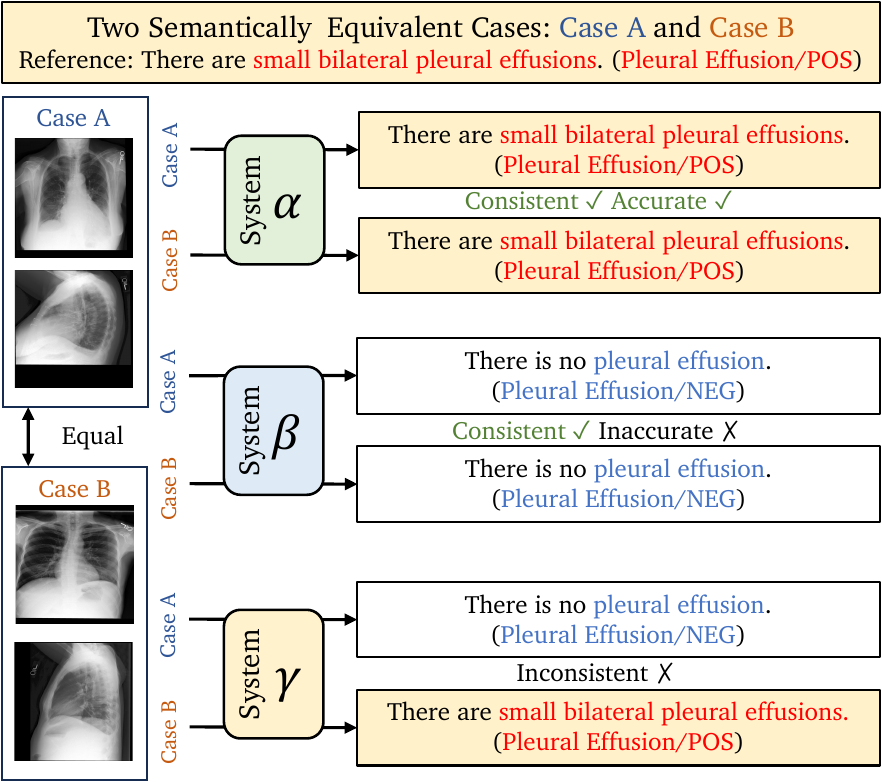}
	\caption{Given two semantically equivalent cases (i.e., Case A and Case B), an example to illustrate the difference between three radiology report generation systems: a consistent and accurate system (i.e., System $\alpha$) and a consistently inaccurate system (i.e., System $\beta$), and an inconsistent system (i.e., System $\gamma$).}
    \label{figure1}
\end{figure}
\section{Introduction}

Being part of the diagnostic process, radiology report generation \cite{7780643,8099861,coatt} has garnered significant attention within the research community, due to its large potential to alleviate the heavy strain of radiologists. Recent research \cite{coplan,rgrg,organ} has made noteworthy progress in enhancing the clinical accuracy of the generated reports.

However, constructing a credible report generation system goes beyond the overall accuracy. There is another crucial quality for report generation systems that has been largely overlooked in the existing literature of medical report generation, which is, \textit{inter-report consistency} \cite{elazar-etal-2021-measuring}. To illustrate the disparity between accuracy and inter-report consistency, we exemplify two semantically equivalent cases as shown in Figure \ref{figure1}, where they share similar observations and reports. Specifically, System $\alpha$ demonstrates the ability to maintain both inter-report consistency and factual accuracy for two similar cases (i.e., \textit{"small bilateral pleural effusions"} for positive \textit{Pleural Effusion}), whereas other systems (i.e., $\beta$ and $\gamma$) fail to meet these criteria. These systems might have overfitted to ordinary cases and could be vulnerable to noise or attack. In terms of enhancing the system’s credibility, inter-report consistency might even hold greater significance than the overall accuracy, since a system prone to providing conflicting results would severely undermine users’ trust \cite{qayyum2020secure,trust_in_ai}. Regrettably, existing report generation systems struggle to maintain this important quality. They tend to exhibit biases towards common patterns, primarily describing normal observations and are susceptible to lesion variants and context noise \cite{r2gen,cmm-rl,attack1,attack2}. We argue that this is largely due to their limited capability of capturing shared attributes of similar patterns, which arises from the data scarcity of distributed lesions and their semantically equivalent variants, rendering it challenging for neural models to accurately locate and describe abnormalities.

In this paper, we propose \textsc{ICon}, which aims to \underline{\textsc{I}}mproves inter-report \underline{\textsc{Con}}sistency of radiology report generation. Our proposed method involves first extracting lesions from given input images, followed by examining the attributes of these lesions. Subsequently, both the radiographs and their associated attributes are utilized as inputs for report generation. To further enhance the inter-report consistency, we introduce a lesion-aware mixup technique by learning from linearly combined lesions and synthesized attributes that belong to the same observation. In summary, the contributions of this paper are as follows:
\begin{itemize}
\item To the best of our knowledge, we are the first to introduce \textit{inter-report consistency} in radiology report generation. To this end, we devise two metrics (\textsc{Con} and \textsc{R-Con}) to measure such consistency.
\item We propose \textsc{ICon}, which improves both the \textit{consistency} and \textit{accuracy} in radiology report generation by capturing abnormalities at the region level. \textsc{ICon} only requires coarse-grained labels (i.e., image labels) for training to extract lesions\footnote{In this context, the term "lesion" generally refers to a specific abnormality. It encompasses most observation categories, excluding \textit{Support Devices}, \textit{Cardiomegaly}, and \textit{Enlarged Cardiomediastinum}. For simplicity, we consider all corresponding regions as lesions.
}, in contrast to previous methods that require fine-grained labels (i.e., bounding boxes).
\item Extensive experiments are conducted on three publicly available datasets, and the results demonstrate the effectiveness of \textsc{ICon} in terms of improving both the consistency and accuracy of the generated reports.
\end{itemize}
\section{Preliminaries}
\subsection{Problem Formulation}
Given a set of radiographs $\mathcal{X} = \{X_1, \dots, X_L\}$ in one study, along with its historical records $\mathcal{X}^p = \{X_1^p, \dots, X_{|p|}^p\}$ (or $\mathcal{X}^p = \emptyset$ if no historical records are available), and its report $\mathcal{Y}$, the task of radiology report generation (RRG) is to generate the report $\mathcal{Y}$ based on $\mathcal{X}$ and $\mathcal{X}^p$. We elaborate on the justification for using historical records as context in Appendix \ref{appendix:justification}. Our proposed method, denoted as \textsc{ICon}, decomposes the RRG task into two stages: Lesion Extraction (Stage 1) and Report Generation (Stage 2). Specifically, given the input images $\mathcal{X}$, \textsc{ICon} first extracts $M$ region candidates $\mathcal{R} = \{R_1, \dots, R_M\}$ and then classifies regions as lesions $\mathcal{Z} = \{Z_1, \dots, Z_{|O|}\}$, where $O$ denotes the observations. Subsequently, in Stage 2, \textsc{ICon} generates a report based on the input images $\mathcal{X}$, historical records $\mathcal{X}^p$, and the extracted lesions $\mathcal{Z}$.

\subsection{Observation and Attribute Annotation}\label{observation_annotation}
\textbf{Observations for Lesion Extraction.} Lesion extraction requires report-level labels, and we adopt CheXbert \cite{chexbert} for this purpose. Specifically, CheXbert annotates a report with 14 observation categories $O=\{o_1, \dots, o_{14}\}$, with data statistics provided in Appendix \ref{appendix: obs_stat}. Each observation is assigned one of four statuses: \textit{Present}, \textit{Absent}, \textit{Uncertain}, and \textit{Blank}. During training and evaluation, \textit{Present} and \textit{Uncertain} are merged into the \textit{Positive} category, which represents abnormal observations. Note that for the observation \textit{No Finding}, only two statuses, \textit{Present} or \textit{Absent}, are applicable. Finally, observation information is utilized for lesion extraction as described in \S\ref{zoomer}.

\noindent\textbf{Attributes for Lesion-Attribute Alignment.} After extracting observations, we further extract entities that represent their characteristics. Specifically, we adopt the attributes released by \citet{recap}\footnote{The attributes are available at \url{https://github.com/wjhou/Recap}.}, which are entities (with a relation \textit{modify} or \textit{located\_at}) extracted from RadGraph \cite{radgraph} using PMI \cite{pmi}. We select the top 30 attributes for each observation and list some of them in Appendix \ref{appendix: attribute} for a better understanding. These attributes are then utilized for lesion-attribute alignment as described in \S\ref{inspector}.

\begin{figure*}[t]
    \centering
    \setlength\belowcaptionskip{\fmargin}
    \includegraphics[width=1.0\linewidth]{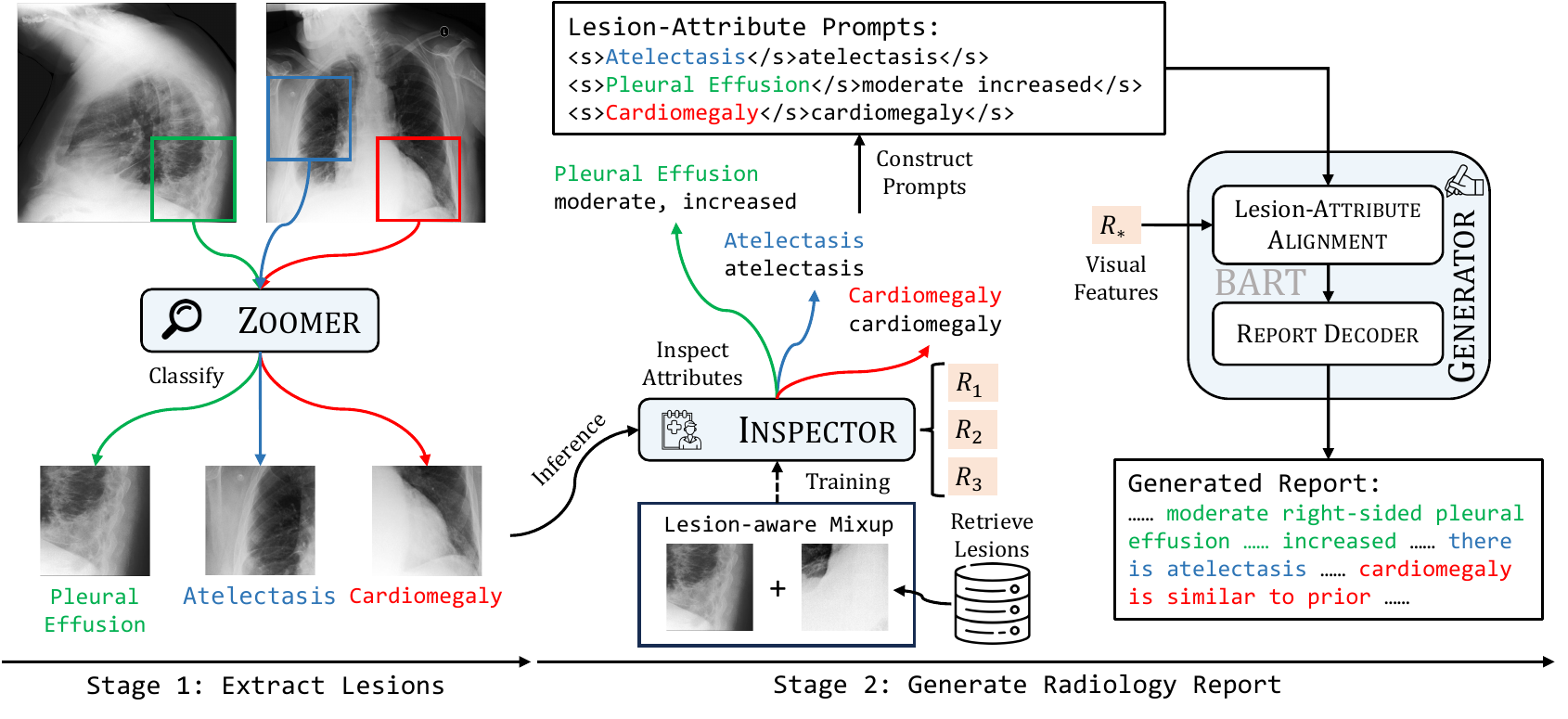}
    \caption{Overview of the \textsc{ICon} framework, which first extracts lesions and then generates reports. Attributes are extracted from RadGraph \cite{radgraph}.}
    \label{figure2}
\end{figure*}
\subsection{Inter-Report Consistency Metrics}\label{CON}
To assess the inter-report consistency of a model, we introduce two metrics, \textsc{Con} and \textsc{R-Con}, inspired by \citet{elazar-etal-2021-measuring}. Semantically equivalent samples should have high observation and entity similarity, which we calculate using the Overlap Coefficient \cite{overlap}: $\mathtt{Overlap}(A,B) = \frac{|A \cap B|}{\mathtt{min}(|A|, |B|)}$. For a report $Q_i$ and its relevant reports $\mathcal{K}_i = \{K_{i,1}, \dots, K_{i,N}\}$, when the observation similarity satisfies $\mathtt{Overlap}(O_{Q_i}, O_{K_{i,j}}) \geq 0.75$ and the entity similarity satisfies $\mathtt{Overlap}(Q_i, K_{i,j}) \geq 0.5$, we regard them as semantically equivalent samples. We then collect the corresponding outputs of $\mathcal{K}_i$ from a model, denoted as $\widehat{\mathcal{K}}_i = \{\widehat{K}_{i,1}, \dots, \widehat{K}_{i,N}\}$. The similarity between two outputs $\widehat{Q}_i$ and $\widehat{K}_{i,j}$ is:
\begin{equation*}
\setlength{\belowdisplayskip}{\margin}
\setlength{\abovedisplayskip}{\margin}
    \mathtt{Overlap}(\widehat{Q}_i, \widehat{K}_{i,j}) = \frac{|\widehat{e}_i \cap \widehat{e}_j|}{\mathtt{min}(|\widehat{e}_i|, |\widehat{e}_j|)},
\end{equation*}
where $\widehat{e}_i$ and $\widehat{e}_j$ are entities and attributes in $\widehat{Q}_i$ and $\widehat{K}_{i,j}$, respectively. The inter-report consistency is then defined as:
\begin{equation*}
\setlength{\belowdisplayskip}{\margin}
\setlength{\abovedisplayskip}{\margin}
    \textsc{Con}(\widehat{Q}_i, \widehat{\mathcal{K}}_i) = \frac{1}{N} \sum_{j=1}^{N} \mathtt{Overlap}(\widehat{Q}_i, \widehat{K}_{i,j}).
\end{equation*}
Since \textsc{Con} only considers inter-report consistency without accounting for reference quality, we introduce \textsc{R-Con}, which considers both consistency and accuracy:
\begin{equation*}
\setlength{\belowdisplayskip}{\margin}
\setlength{\abovedisplayskip}{\margin}
    \textsc{R-Con}(\widehat{Q}_i, \widehat{\mathcal{K}}_i) = \tau_i \cdot \textsc{Con}(\widehat{Q}_i, \widehat{\mathcal{K}}_i),
\end{equation*}
where $\tau_i = \mathtt{Overlap}(\widehat{Q}_i, Q_i)$ is the similarity between the hypothesis and its reference.
\section{Methodology}\label{section3}
\subsection{Visual Encoding}\label{visual_encoding}
Given an image $X_l$, an image processor is first utilized to split $X_l$ into $N$ patches. Then, a visual encoder $f_{\theta}$, e.g., Swin Transformer \cite{swintransformer},  is employed to extract visual representations $\bm{X}_l$ and the pooler output $\bm{P}_l \in \mathbb{R}^h$:
\begin{equation*}
\setlength{\belowdisplayskip}{\margin}
\setlength{\abovedisplayskip}{\margin}
    \begin{split}
        [\bm{P}_l, \bm{X}_l] = f_{\theta}(X_l),
    \end{split}
\end{equation*}
where $\bm{X}_l = \{\bm{x}_{l,i}, \dots, \bm{x}_{l,N}\}$ and $\bm{x}_{l,i} \in \mathbb{R}^h$ is the $i$-th visual representation.

\subsection{Stage 1: Extracting Lesions via Observation Classification (\textsc{Zoomer})}\label{zoomer}
\textbf{Observation Classification.} A \textsc{Zoomer} is a visual encoder parameterized by $\theta_\text{Z}$ and trained to classify a given input $\mathcal{X}$ into abnormal observations as mentioned in \S\ref{observation_annotation}:
\begin{equation*}
\setlength{\belowdisplayskip}{\margin}
\setlength{\abovedisplayskip}{\margin}
    p(o_i)=\textsc{Zoomer}(\mathcal{X}).
\end{equation*}
Specifically, \textsc{Zoomer} first encodes images $\mathcal{X}=\{X_1,\dots,X_L\}$ as outlined in \S\ref{visual_encoding}, and then takes the averaged pooler output for classification, following these steps:
\begin{equation*}
\setlength{\belowdisplayskip}{\margin}
\setlength{\abovedisplayskip}{\margin}
    \begin{gathered}
    [\bm{P}_l, \bm{X}_l] = f_{\theta_\text{Z}}(X_l), \\
    \bm{P} = \frac{1}{L}\sum \bm{P}_l, \\
    p(o_i) = \sigma(\bm{W}_i \bm{P} + b_i),
    \end{gathered}
\end{equation*}
where $\bm{W}_i \in \mathbb{R}^h$ is the weight for the $i$-th observation, $b_i \in \mathbb{R}$ is its bias, and $\sigma$ is the Sigmoid function.

\noindent\textbf{Zooming In for Lesion Extraction.} Upon completing training \textsc{Zoomer}, we can use it to extract lesions without the need for object detectors \cite{faster_rcnn}. It is worth noting that our method does not require fine-grained labels, such as bounding boxes \cite{rgrg}.

For an image $X_l$, a sliding window with a $0.375$ ratio of $X_l$ is applied to extract $M$ region candidates $\mathcal{R}_{l}=\{R_{l,1}, \dots, R_{l,M}\}$ from $X_l$, as shown in the left side of Figure \ref{figure2}. These regions are then sequentially fed into \textsc{Zoomer} for classification. Further details on the extraction of these regions can be found in Appendix \ref{appendix: extract_region}. The probability of a region $R_{l,j}$ being classified as an abnormal observation $o_i$ is:
\begin{equation*}
\setlength{\belowdisplayskip}{\margin}
\setlength{\abovedisplayskip}{\margin}
    p_{l,j}(o_i) = \textsc{Zoomer}(R_{l,j}).
\end{equation*}
For each study, all images in $\mathcal{X}$ are iterated, and only the region with the highest $p_{l,j}(o_i)$ is chosen as a lesion $Z_i$ corresponding to the observation $o_i$. Finally, the set of lesions is denoted as $\mathcal{Z}=\{Z_1, \dots, Z_{|O|}\}$.

\noindent\textbf{Training \textsc{Zoomer}.} \textsc{Zoomer} is optimized using the binary cross-entropy (BCE) loss. To handle the class-imbalanced issue (refer to Appendix \ref{appendix: obs_stat} for details), a weight factor $\alpha_j$ is applied for each abnormal observation, and the loss function $\mathcal{L}_{\text{S1}}$ is:
\begin{equation*}
\setlength{\belowdisplayskip}{\margin}
\setlength{\abovedisplayskip}{\margin}
    \begin{split}
        \mathtt{BCE}(p(o_j), o_j) &= -\frac{1}{|O|} \sum_{j} \left[ \alpha_j \cdot o_j \cdot \mathtt{log} \: p(o_j) \right. \\
        & \quad \left. + (1 - o_j) \cdot \mathtt{log}(1 - p(o_j)) \right],
    \end{split}
\end{equation*}
where $o_j \in \{0, 1\}$ is the label, $\alpha_j = 1 + \mathtt{log} \left( \frac{|\mathcal{D}_{\text{train}}| - w_j}{w_j} \right)$, and $|\mathcal{D}_{\text{train}}|$ and $w_j$ are the number of samples and the number of $j$-th observations in the training set, respectively.

\begin{table*}[t]
    \centering
    \setlength\belowcaptionskip{\fmargin}
    \resizebox{\textwidth}{!}{
    \begin{tabular}{c|l|cccccc|ccc}
    \Xhline{2\arrayrulewidth}
    \multirow{2}{*}{\textbf{Dataset}} & \multirow{2}{*}{\textbf{Model}} & \multicolumn{6}{c|}{\textbf{NLG Metrics}} & \multicolumn{3}{c}{\textbf{CE Metrics}} \\
    \cline{3-11}
    & & \textbf{B-1} & \textbf{B-2} & \textbf{B-3} & \textbf{B-4} & \textbf{MTR} & \textbf{R-L} & \textbf{P} & \textbf{R} & \textbf{F}$_1$ \\ 
    \hline
    \multirow{5}{*}{\textsc{\makecell{MIMIC \\ -ABN}}} & \textsc{R2Gen} & $0.290$ & $0.157$ & $0.093$ & $0.061$ & $0.105$ & $0.208$ & $0.266$ & {$0.320$} & {$0.272$} \\
    & \textsc{R2GenCMN} & $0.264$ & $0.140$ & $0.085$ & $0.056$ & $0.098$ & $0.212$ & {$0.290$} & $0.319$ & $0.280$ \\
    & \textsc{ORGan} & {$0.314$} & {$0.180$} & {$0.114$} & {$0.078$} & \underline{$0.120$} & \underline{$0.234$} & $0.271$ & {$0.342$} & {$0.293$} \\
    &\textsc{Recap} & \underline{$0.321$} & \underline{$0.182$} & \underline{$0.116$} & \underline{$0.080$} & \underline{$0.120$} & {$0.223$} & \underline{$0.300$} & \underline{$0.363$} & \underline{$0.305$}\\
    &\textsc{ICon} (Ours) & $\bm{0.337}$ & $\bm{0.195}$ & $\bm{0.126}$ & $\bm{0.086}$ & $\bm{0.129}$ & $\bm{0.236}$ & $\bm{0.332}$ & $\bm{0.430}$ & $\bm{0.360}$\\
    \Xhline{2\arrayrulewidth}
    \multirow{13}{*}{\textsc{\makecell{MIMIC \\ -CXR}}}
    & \textsc{R2Gen} & $0.353$ & $0.218$ & $0.145$ & $0.103$ & $0.142$ & $0.270$ & $0.333$ & $0.273$ & $0.276$ \\
    & \textsc{R2GenCMN} & $0.353$ & $0.218$ & $0.148$ & $0.106$ & $0.142$ & $0.278$ & $0.344$ & $0.275$ & $0.278$ \\
    & $\mathcal{M}^2$\textsc{Tr} & $0.378$ & $0.232$ & $0.154$ & $0.107$ & $0.145$ & {$0.272$} & $0.240$ & {$0.428$} & $0.308$ \\
    & \textsc{KnowMat} & $0.363$ & $0.228$ & $0.156$ & $0.115$ & $-$ & $0.284$ & \bm{$0.458$} & $0.348$ & $0.371$ \\
    & \textsc{CMM-RL} & {$0.381$} & $0.232$ & $0.155$ & $0.109$ & {$0.151$} & {$0.287$} & $0.342$ & $0.294$ & $0.292$ \\
    & \textsc{CMCA} & $0.360$ & $0.227$ & {$0.156$} & $0.117$ & $0.148$ & {$0.287$} & {${0.444}$} & $0.297$ & $0.356$ \\
    & KiUT & {$0.393$} & $0.243$ & $0.159$ & $0.113$ & {$0.160$} & $0.285$ & $0.371$ & $0.318$ & $0.321$ \\
    & DCL & $-$ & $-$ & $-$ & $0.109$ & $0.150$ & $0.284$ & $0.471$ & $0.352$ & {$0.373$} \\
    & METrans & $0.386$ & {$0.250$} & {$0.169$} & {$0.124$} & $0.152$ & \underline{$0.291$} & $0.364$ & $0.309$ & $0.311$ \\
    & RGRG & $0.373$ & $0.249$ & $0.175$ & $\bm{0.126}$ & $0.168$ & $0.264$ & $0.380$ & $0.319$ & $0.305$ \\
    & \textsc{ORGan} & {$0.407$} & {$0.256$} & {$0.172$} & $0.123$ & {$0.162$} & $\bm{0.293}$ & $0.416$ & $0.418$ & {$0.385$} \\
    &\textsc{Recap} & \bm{$0.429$} & $\bm{0.267}$ & $\underline{0.177}$ & ${0.125}$ & $\underline{0.168}$ & $0.288$ & $0.389$ & $\underline{0.443}$ & $\underline{0.393}$ \\
    &\textsc{ICon} (Ours) & $\bm{0.429}$ & $\underline{0.266}$ & $\bm{0.178}$ & $\bm{0.126}$ & $\bm{0.170}$ & $0.287$ & $\underline{0.445}$ & $\bm{0.505}$ & $\bm{0.464}$ \\
    \cline{2-11}
    \Xhline{2\arrayrulewidth}
    \end{tabular}}
    \caption{Experimental results of our model and baselines on the \textsc{MIMIC-ABN} and \textsc{MIMIC-CXR} datasets. The best results are in \textbf{boldface}, and the \underline{underlined} are the second-best results.}
    \label{table: mimic_results}
\end{table*}

\subsection{Stage 2: Inspecting Lesions (\textsc{Inspector})}\label{inspector}
\textbf{Inspecting Lesions with Attributes.} Given that lesions of the same observation can exhibit different characteristics, it is crucial to inspect each lesion and match it with corresponding attributes (\S\ref{observation_annotation}) to differentiate it from other variations. Specifically, an \textsc{Inspector} is a visual encoder parameterized by $\theta_{I}$, similar to \S\ref{zoomer}. \textsc{Inspector}($\bm{{P}}^p, \bm{{P}}, Z_j$) takes prior and current visit chest X-rays as context, along with a lesion region as input:
\begin{equation*}
\setlength{\belowdisplayskip}{\margin}
\setlength{\abovedisplayskip}{\margin}
    \begin{gathered}
        [\bm{P}_{Z_j}, \bm{Z}_j] = f_{\theta_I}(Z_j), \\
        p_{j}(a_{k}) = \sigma(\mathtt{MLP}(\bm{{P}}^p, \bm{{P}}, \bm{P}_{Z_j})),
    \end{gathered}
\end{equation*}
where $\mathtt{MLP}$ is a two-layer perceptron with non-linear activation, and $\bm{{P}}^p, \bm{{P}}, \bm{P}_{Z_j} \in \mathbb{R}^h$ are pooler outputs of prior images, current images, and the lesion, respectively. Concurrently, the lesion features $\bm{\mathcal{Z}} = \{\bm{Z}_1, \dots, \bm{Z}_{|O|}\}$ are collected for report generation. For image encoding, we use another visual encoder $f_{\theta_V}$ to encode $\mathcal{X}$ into $\bm{\mathcal{X}}$ and $\mathcal{X}^p$ into $\bm{\mathcal{X}}^p$. By inspecting lesion-level features, \textsc{ICon} can capture fine-grained details which are beneficial for generating consistent outputs.

\begin{figure}[t]
    \centering
    \setlength\belowcaptionskip{\fmargin}
    \includegraphics[width=1.0\linewidth]{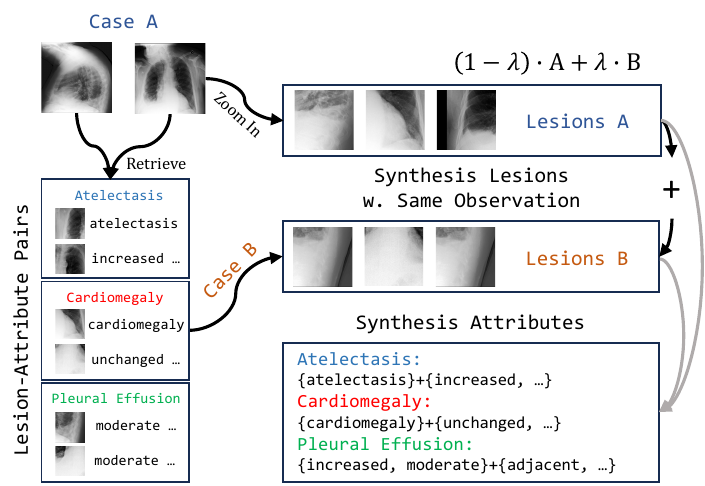}
    \caption{Overview of our proposed lesion-aware mixup augmentation.}
    \label{figure: mixup}
\end{figure}

\noindent\textbf{Lesion-aware Mixup.} To further improve the consistency of the generated outputs, we adopt the mixup augmentation method \cite{zhang2018mixup} and devise a lesion-aware mixup during the training phase. Specifically, for a lesion-attribute pair $(Z_j, A_j)$, we retrieve a similar pair $(Z_k, A_k)$ with the same observation from the training data based on report similarity. These lesions are synthesized by a linear combination, as illustrated in Figure \ref{figure: mixup}:
\begin{equation*}
\setlength{\belowdisplayskip}{\margin}
\setlength{\abovedisplayskip}{\margin}
    Z^*_{j} = \lambda Z_{j} + (1 - \lambda) Z_{k},
\end{equation*}
where $\lambda$ is set to 0.75. Note that during training, $Z^*_j$ is used for both \textsc{Inspector} and \textsc{Generator}.

\noindent\textbf{Training \textsc{Inspector}.}
Similar to \S\ref{zoomer}, we adopt a linearly combined BCE loss to optimize \textsc{Inspector}:
\begin{equation*}
\setlength{\belowdisplayskip}{\margin}
\setlength{\abovedisplayskip}{\margin}
    \mathcal{L}_\text{I}=\lambda\mathtt{BCE}_j + (1 - \lambda)\mathtt{BCE}_k,
\end{equation*}
where $\mathtt{BCE}_j$ and $\mathtt{BCE}_k$ take $A_j$ and $A_k$ as their respective labels. Notably, only the attributes that are shared between $Z_j$ and $Z_k$ are fully optimized. Consequently, our lesion-aware mixup technique facilitates the improvement of output consistency for two semantically equivalent lesions.

\subsection{Stage 2: Generating Reports (\textsc{Generator})}
\textbf{Lesion-Attribute Alignment.} To bridge the modality gap between lesion representations and attributes, we leverage a BART \cite{bart} encoder to extract attribute representations. The attributes associated with each lesion are formulated as a prompt: \texttt{<s>} $o_j$ \texttt{</s>} $A_j$ \texttt{</s>}, as depicted in Figure \ref{figure2}. Then, a cross-attention module \cite{Transformer} is inserted after every self-attention module. This module aligns the lesion representations with the attribute representations by querying visual representations using attribute representations, similar to Q-Former \cite{li2023blip2}:
\begin{equation*}
\setlength{\belowdisplayskip}{\margin}
\setlength{\abovedisplayskip}{\margin}
    \bm{H}^a_j = \mathtt{CrossAttention}(\bm{H}^s_j, \bm{Z}_j, \bm{Z}_j),
\end{equation*}
where $\bm{H}^a_j, \bm{H}^s_j \in \mathbb{R}^h$ are the aligned attribute representation and the self-attended representation of $A_j$, respectively. All prompts are encoded, and the attribute representations of $\bm{\mathcal{Z}}$ are denoted as $\bm{\mathcal{H}}^a$.

\noindent\textbf{Report Generation.} Given the input images $\bm{\mathcal{X}}$, images of prior visits $\bm{\mathcal{X}}^p$, the lesions $\bm{\mathcal{Z}}$, and attribute $\bm{\mathcal{H}}^a$, we utilize a BART decoder in conjunction with the Fusion-in-Decoder (FiD; \cite{izacard-grave-2021-leveraging}) that simply concatenates multiple context sequences for report generation. Then, the probability of the $t$-th step is expressed as:
\begin{equation*}
\setlength{\belowdisplayskip}{\margin}
\setlength{\abovedisplayskip}{\margin}
    \begin{gathered}
        \bm{h}_t = \mathtt{FiD}([\bm{\mathcal{X}}; \bm{\mathcal{X}}^p; \bm{Z}; \bm{\mathcal{H}}^a], \bm{h}_{< t}), \\
        p(y_t | \mathcal{X}, \mathcal{X}^p, \mathcal{Z}, \mathcal{Y}_{< t}) = \mathtt{Softmax}(\bm{W}_g \bm{h}_t + \bm{b}_g),
    \end{gathered}
\end{equation*}
where $\bm{h}_t \in \mathbb{R}^h$ is the $t$-th hidden representation, $\bm{W}_g \in \mathbb{R}^{|\mathcal{V}| \times h}$ is the weight matrix, $\bm{b}_g \in \mathbb{R}^{|\mathcal{V}|}$ is the bias vector, and $\mathcal{V}$ is the vocabulary.

\noindent\textbf{Training \textsc{Generator}.} The generation process is optimized using the negative log-likelihood loss, given each token's probability $p(y_t | \mathcal{X}, \mathcal{X}^p, \mathcal{Z}, \mathcal{Y}_{< t})$:
\begin{equation*}
\setlength{\belowdisplayskip}{\margin}
\setlength{\abovedisplayskip}{\margin}
    \begin{split}
        \mathcal{L}_{\text{G}} = -\sum^{T}_{t=1} \mathtt{log} \: p(y_t | \mathcal{X}, \mathcal{X}^p, \mathcal{Z}, \mathcal{Y}_{< t}).
    \end{split}
\end{equation*}
The loss function of Stage 2 is: $\mathcal{L}_{\text{S2}} = \mathcal{L}_{\text{I}} + \mathcal{L}_{\text{G}}$.

\section{Experiments}
\subsection{Datasets}
Three public datasets are used to evaluate our models, i.e., \textsc{IU X-ray}\footnote{\url{https://openi.nlm.nih.gov/}} \cite{iu_xray}, \textsc{MIMIC-CXR}\footnote{\url{https://physionet.org/content/mimic-cxr-jpg/2.0.0/}} \cite{mimic_cxr}, and \textsc{MIMIC-ABN} \cite{mimic_abn}. We follow previous research \cite{r2gen} to preprocess these datasets, and provide other details in Appendix \ref{appendix: data_preprocessing}.
\begin{itemize}[noitemsep,topsep=2pt]
    \item \textsc{IU X-ray} consists of 3,955 reports. We follow previous research \cite{r2gen} and split the dataset into train/validation/test sets with a ratio of 7:1:2.
    \item \textsc{MIMIC-CXR} consists of 377,110 chest X-ray images and 227,827 reports.
    \item \textsc{MIMIC-ABN} is modified from the \textsc{MIMIC-CXR} dataset and its reports only contain abnormal part. We adopt the data-split as used in \citet{recap}, and the data-split is 71,786/546/806 for train/validation/test sets.
\end{itemize}
Unlike previous research \cite{r2gen} which only used one view for report generation on \textsc{MIMIC-CXR} and \textsc{MIMIC-ABN}, we collect all views for each visit in experiments. The justification is provided in Appendix \ref{appendix:justification}.
\begin{table}[t]
    \centering
    \setlength\belowcaptionskip{\fmargin}
    \resizebox{\linewidth}{!}{
    \begin{tabular}{l|l|cc|ccc}
    \Xhline{2\arrayrulewidth}
    \multirow{2}{*}{\textbf{Dataset}} & \multirow{2}{*}{\textbf{Model}} & \multicolumn{2}{c|}{\textbf{NLG Metrics}} & \multicolumn{3}{c}{\textbf{RadGraph}} \\
    \cline{3-7}
    & & \textbf{B-4} & \textbf{R-L} & $\textbf{RG}_\text{E}$ & $\textbf{RG}_\text{ER}$ &$\textbf{RG}_{\overline{\text{ER}}}$ \\\hline
    \multirow{4}{*}{\textsc{\makecell{IU \\ X-ray}}} & \textsc{R2Gen} & ${0.120}$ & $0.298$ & $-$ & $-$ & $-$ \\
    & $\mathcal{M}^2$\textsc{Tr} & $\bm{0.121}$ & $0.288$ & $-$ & $-$ & $-$ \\
    & $\mathcal{T}_{\text{NLL}}$ & $0.114$ & $-$ & $0.230$ & $0.202$ & $0.153$  \\
    &\textsc{ICon} & $0.098$ & $\bm{0.320}$ & $\bm{0.342}$ & $\bm{0.312}$ & $\bm{0.246}$ \\
    \hline
    \multirow{4}{*}{\textsc{\makecell{MIMIC \\ -CXR}}}&$\mathcal{T}_{\text{NLL}}$ & $0.105$ & $0.253$ & $0.230$ & $0.202$ & $0.153$  \\
    & \textsc{ORGan} & $0.123$ & $\bm{0.293}$ & $0.303$ & $0.275$ & $0.199$ \\
    & \textsc{Recap} & $0.125$ & $0.288$ & $0.307$ & $0.276$ & $\bm{0.205}$ \\
    &\textsc{ICon} & $\bm{0.126}$ & $0.287$ & $\bm{0.312}$ & $\bm{0.278}$ & $0.197$ \\
    \Xhline{2\arrayrulewidth}
    \end{tabular}
    }
    \caption{Radgraph evaluation results on the \textsc{IU X-ray} and \textsc{MIMIC-CXR} datasets. Results of $\mathcal{T}_{\text{NLL}}$ are cited from \citet{delbrouck-etal-2022-improving}.}
    \label{table: radgraph_results}
\end{table}
\subsection{Evaluation Metrics and Baselines}
\textbf{NLG Metrics.} To assess the quality of generated reports, we adopt several natural language generation (NLG) metrics for evaluation. BLEU \cite{bleu}, METEOR \cite{meteor}, and ROUGE \cite{rouge} are selected as NLG Metrics, and we use the MS-COCO caption evaluation tool\footnote{\url{https://github.com/tylin/coco-caption}} to compute the results. 

\noindent\textbf{CE Metrics.} Following previous research \cite{r2gen,r2gencmn}, we adopt clinical efficacy (CE) metrics to evaluate the observation-level factual accuracy, and CheXbert \cite{chexbert} is used in this paper. To measure the entity-level factual accuracy, we leverage the RadGraph \cite{radgraph,delbrouck-etal-2022-improving} and temporal entity matching (TEM) scores proposed by \citet{bannur2023learning} for evaluation.

\noindent\textbf{Consistency Metrics.} \textsc{Con} and \textsc{R-Con} (\S\ref{CON}) are utilized to measure the inter-report consistency. Note that entities used in measuring consistency are adopted from RadGraph \cite{radgraph}. A \textsc{Majority} baseline which outputs the same report for all inputs, is included.

\noindent\textbf{Baselines.} We compare our models with the following state-of-the-art (SOTA) baselines: \textsc{R2Gen} \cite{r2gen}, \textsc{R2GenCMN} \cite{r2gencmn}, \textsc{KnowMat} \cite{mia}, $\mathcal{M}^2$\textsc{Tr} \cite{m2tr}, \textsc{CMM-RL} \cite{cmm-rl}, \textsc{CMCA} \cite{cmca}, {CXR-RePaiR-Sel/2} \cite{cxr_repair_sel}, {BioViL-T} \cite{bannur2023learning}, DCL \cite{dcl}, METrans \cite{metrans}, KiUT \cite{kiut}, RGRG \cite{rgrg}, \textsc{ORGan} \cite{organ}, and \textsc{Recap} \cite{recap}.
\begin{table}[t]
    \centering
    \setlength\belowcaptionskip{\fmargin}
    \resizebox{\linewidth}{!}{
    \begin{tabular}{l|cc|cc}
    \Xhline{2\arrayrulewidth}
    \multirow{2}{*}{\textbf{Model}} & \multicolumn{2}{c|}{{\textbf{MIMIC-ABN}}} &\multicolumn{2}{c}{{\textbf{MIMIC-CXR}}} \\
    \cline{2-5}
    & \textbf{\textsc{Con}} & \textbf{\textsc{R-Con}} & \textbf{\textsc{Con}} & \textbf{\textsc{R-Con}}\\ 
    \hline
    \textsc{Majority} & $1.000$ & $-$ & $1.000$ & $-$ \\ \hline
    \textsc{R2Gen} & $0.280$ & $0.072$ & $0.137$ & $0.042$ \\
    \textsc{R2GenCMN} & $0.302$ & $0.091$ & $0.155$ & $0.049$\\
    \textsc{ORGan} & $\bm{0.338}$ & $0.127$ & $0.345$ & $0.126$ \\
    \textsc{Recap} & $0.311$ & $0.108$ & $0.345$ & $0.114$ \\
    \Xhline{2\arrayrulewidth}
    \textsc{ICon} (Ours) & $0.316$ & $\bm{0.140}$ & $\bm{0.351}$ & $\bm{0.163}$ \\
    \textsc{ICon} \textit{w/o} \textsc{Zoom} & $0.183$ & $0.073$ & $0.175$ & $0.066$ \\
    \textsc{ICon} \textit{w/o} \textsc{Inspect} & $0.253$ & $0.100$ & $0.245$ & $0.090$ \\
    \textsc{ICon} \textit{w/o} \textsc{Mixup} & $0.286$ & $0.119$ & $0.334$ & $0.156$\\
    \Xhline{2\arrayrulewidth}
    \end{tabular}}
    \caption{The \textsc{Con} score and the \textsc{R-Con} score. \textsc{Majority}: outputs the same report for all inputs.}
    \label{table: consistency_results}
\end{table}
\subsection{Implementation Details}
The small and tiny versions of Swin Transformer V2 \cite{swintransformer_v2} are used as the visual backbone for \textsc{Zoomer} and \textsc{Inspector}, respectively. The \textsc{Generator} is initialized with the base version of BART \cite{bart} pretrained on biomedical corpus \cite{biobart}. Other parameters are randomly initialized. For Stage 2 training, the learning rate is $5e-5$ with linear decay, the batch size is $32$, and the models are trained for $20$ and $5$ epochs on \textsc{MIMIC-ABN} and \textsc{MIMIC-CXR} with early stopping, respectively. Since the number of samples in \textsc{IU X-ray} is too small to train a multimodal model, we only provide results produced by models trained on \textsc{MIMIC-CXR} as a reference, similar to \cite{delbrouck-etal-2022-improving}. For other training details, and the resources used in this paper, we list them in Appendix \ref{appendix: s1_impl}.
\begin{table*}[t]
    \centering
    \setlength\belowcaptionskip{\fmargin}
    \resizebox{\textwidth}{!}{
    \begin{tabular}{c|l|ccc|cccccc|ccc}
    \Xhline{2\arrayrulewidth}
    \multirow{2}{*}{\textbf{Dataset}} & \multirow{2}{*}{\textbf{Model}} & \multicolumn{3}{c|}{\textbf{Components}} & \multicolumn{6}{c|}{\textbf{NLG Metrics}} & \multicolumn{3}{c}{\textbf{CE Metrics}} \\\cline{3-14}
    & & \textbf{\textsc{Zoom}} & \textbf{\textsc{Inspect}} & \textbf{\textsc{Mixup}} & \textbf{B-1} & \textbf{B-2} & \textbf{B-3} & \textbf{B-4} & \textbf{MTR} & \textbf{R-L} & \textbf{P} & \textbf{R} & \textbf{F}$_1$ \\ 
    \hline
    \multirow{4}{*}{\textsc{\makecell{MIMIC\\-ABN}}} & \textsc{ICon} & \Checkmark & \Checkmark & \Checkmark & $0.337$ & $0.195$ & $0.126$ & $0.086$ & $0.129$ & $0.236$ & $0.332$ & $0.430$ & $0.360$ \\
    & \textsc{ICon} \textit{w/o} \textsc{Zoom} & \XSolidBrush & \XSolidBrush & \XSolidBrush & $0.310$ & $0.181$  & $0.119$ & $0.084$ & $0.120$ & $0.243$ & $0.306$ & $0.353$ & $0.306$ \\
    & \textsc{ICon} \textit{w/o} \textsc{Inspect} & \Checkmark & \XSolidBrush & \XSolidBrush & $0.315$ & $0.182$ & $0.117$ & $0.081$ & $0.121$ & $0.236$ & $0.338$ & $0.401$ & $0.352$\\
    & \textsc{ICon} \textit{w/o} \textsc{Mixup} & \Checkmark & \Checkmark & \XSolidBrush & $0.335$ & $0.192$ & $0.124$ & $0.085$ & $0.129$ & $0.239$ & $0.332$ & $0.413$ & $0.356$ \\
    \Xhline{2\arrayrulewidth}
    \multirow{4}{*}{\textsc{\makecell{MIMIC\\-CXR}}} & \textsc{ICon} & \Checkmark & \Checkmark & \Checkmark & ${0.429}$ & ${0.266}$ & ${0.178}$ & ${0.126}$ & ${0.170}$ & $0.287$ & ${0.445}$ & ${0.505}$ & ${0.464}$ \\
    & \textsc{ICon} \textit{w/o} \textsc{Zoom} & \XSolidBrush & \XSolidBrush & \XSolidBrush & $0.377$ & $0.237$ & $0.162$ & $0.119$ & $0.149$ & $0.288$ & $0.363$ & $0.280$ & $0.278$ \\
    & \textsc{ICon} \textit{w/o} \textsc{Inspect} & \Checkmark & \XSolidBrush & \XSolidBrush & $0.399$ & $0.248$ & $0.168$ & $0.122$ & $0.157$ & $0.287$ & $0.444$ & $0.447$ & $0.423$ \\
    & \textsc{ICon} \textit{w/o} \textsc{Mixup} & \Checkmark & \Checkmark & \XSolidBrush & $0.427$ & $0.264$ & $0.176$ & $0.124$ & $0.169$ & $0.285$ & $0.444$ & $0.502$ & $0.462$\\
    \Xhline{2\arrayrulewidth}
    \end{tabular}}
    \caption{Ablation results of our model and its variants on the \textsc{MIMIC-ABN} and \textsc{MIMIC-CXR} datasets. A \Checkmark indicates that the component is included, while an \XSolidBrush denotes that it is removed.}
    \label{table: mimic_ablation}
\end{table*}
\begin{table}[t]
\centering
\setlength\belowcaptionskip{\fmargin}
    \resizebox{\linewidth}{!}
    {
    \begin{tabular}{l|cccc}
    \Xhline{2\arrayrulewidth}
    \textbf{Model} & \textbf{B-4} & \textbf{R-L} & \textbf{CE-F$_1$} & \textbf{TEM} \\\hline
    {CXR-RePaiR-2} & $0.021$ & $0.143$ & $0.281$ & $0.125$  \\
    {BioViL-NN} & $0.037$ & $0.200$ & $0.283$ & $0.111$ \\
    {BioViL-T-NN} & $0.045$ & $0.205$ & $0.290$ & $0.130$ \\
    {BioViL-AR} & $0.075$ & $0.279$ & $0.293$ & $0.138$ \\
    {BioViL-T-AR} & $0.092$ & $\bm{0.296}$ & $0.317$ & $0.175$ \\
    \textsc{Recap} & {$0.118$} & $0.279$ & {$0.400$} & {$0.304$}\\
    \textsc{ICon} (Ours) & $\bm{0.120}$ & $0.279$ & $\bm{0.468}$ & $\bm{0.335}$\\
    \Xhline{2\arrayrulewidth}
    \end{tabular}}
    \caption{Progression modeling results on the \textsc{MIMIC-CXR} dataset. Results of BioViL-* are cited from \citet{bannur2023learning}.}
    \label{table: pro_results}
\end{table}
\section{Results}
\subsection{Quantitative Analysis}
\textbf{Inter-Report Consistency Analysis.} Table \ref{table: consistency_results} provides \textsc{Con} and \textsc{R-Con} scores of baselines, our model, and its ablated variants. {\textsc{ICon} achieves the highest \textsc{R-Con} on both datasets, indicating the best inter-report consistency.} In terms of the \textsc{Con} score, \textsc{ICon} demonstrates competitive performance when compared with the best baseline, i.e., \textsc{ORGan}. We also observe that introducing mixup augmentation leads to a moderate improvement in \textsc{Con}, demonstrating the effectiveness of lesion-aware mixup.

\noindent\textbf{NLG and Temporal Modeling Results.} The NLG results are presented in Table \ref{table: mimic_results} and the temporal modeling results are listed in Table \ref{table: pro_results}. Among all models, {\textsc{ICon} achieves SOTA performance on the NLG and temporal metrics.} As shown in Table \ref{table: mimic_results}, our model demonstrates significant improvements on the \textsc{MIMIC-ABN} dataset and achieves competitive performance on the \textsc{MIMIC-CXR} dataset. Additionally, we provide experimental results on the \textsc{IU X-ray} dataset as a reference in Table \ref{table: radgraph_results}. Regarding temporal modeling, \textsc{ICon} exhibits significant improvements over other baselines in terms of BLEU score, clinical accuracy, and TEM score while maintaining competitive performance on ROUGE, indicating its enhanced capacity to utilize historical records effectively.

\noindent\textbf{Clinical Efficacy Results.} In the right section of Table \ref{table: mimic_results}, we observe that \textsc{ICon} achieves SOTA clinical efficacy, increasing the macro CE $\text{F}_1$ score from $0.393$ to $0.464$ on the \textsc{MIMIC-CXR} dataset and rising by $5.5\%$ on the \textsc{MIMIC-ABN} dataset. These results indicate that our model is capable of generating accurate radiology reports. Furthermore, Table \ref{table: radgraph_results} presents the RadGraph F$_1$ score on both the \textsc{IU X-ray} and \textsc{MIMIC-CXR} datasets. Our model achieves competitive performance compared with the non-RL-optimized baselines. We also provide per-observation CE results in Table \ref{table: s2_obs}, example-based CE results in Table \ref{table: example_based_ce}, and the performance of \textsc{Zoomer} in Table \ref{table: s1_zoomer} for reference.

\begin{figure*}[t]
	\centering
    \setlength\belowcaptionskip{\fmargin}
    \resizebox{\textwidth}{!}{
        \begin{sanseriffont}
            \begin{tabular}{@{}p{0.35\textwidth}@{}}
            \centerline{ \includegraphics[width=1.0\linewidth]{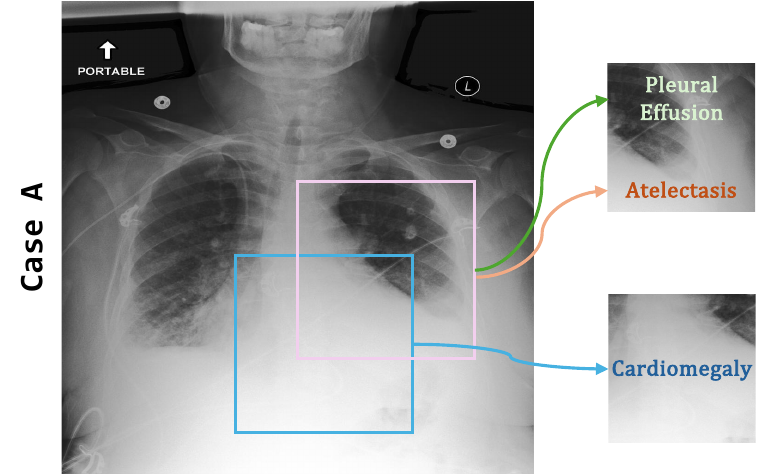}} \\
            \end{tabular}
        \end{sanseriffont}
        \fbox{
            \begin{sanseriffont}
                \begin{tabular}{@{}p{0.65\textwidth}@{}}
                    \textbf{Reference A:} in comparison with the study of there is little overall change. {\hlomma{continued enlargement of the cardiac silhouette}} with \underline{\hlommd{pulmonary vascular congestion}} and \underline{\hlommb{bilateral pleural effusions}} with \underline{\hlommc{compressive atelectasis}}. central catheter remains in place.\\
            \addlinespace
            \textbf{\textsc{ICon} \textit{w/o} \textsc{Zoom}:} as compared to the previous radiograph the patient has received a right-sided picc line. the course of the line is unremarkable the tip of the line projects over the mid svc. there is no evidence of complications notably no pneumothorax. otherwise the radiograph is unchanged. \\
            \addlinespace
            \textbf{\textsc{ICon}:} in comparison with the study of there is little overall change. {\hlomma{continued enlargement of the cardiac silhouette}} with \underline{\hlommd{pulmonary vascular congestion}} and mild to moderate cardiomegaly. small \underline{\hlommb{bilateral pleural effusions}} with areas of \underline{\hlommc{compressive atelectasis}}. the right picc line has been removed. nasogastric tube remains in place. \\
                \end{tabular}
            \end{sanseriffont}
        } 
    }
    \resizebox{\textwidth}{!}{
        \begin{sanseriffont}
            \begin{tabular}{@{}p{0.35\textwidth}@{}}
            \centerline{ \includegraphics[width=1.0\linewidth]{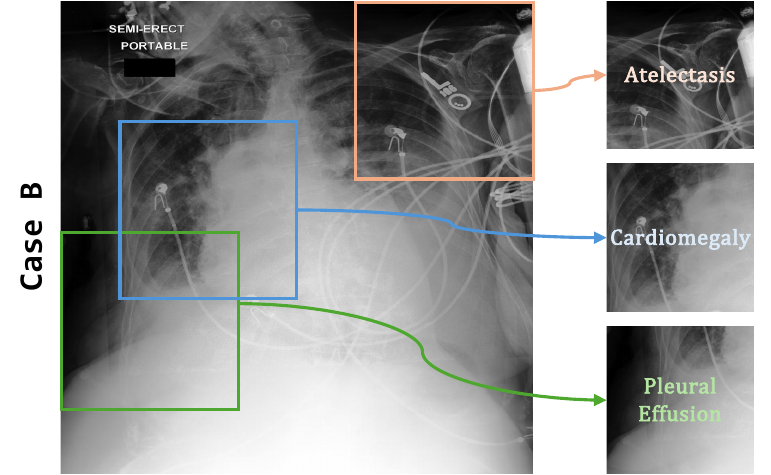}} \\
            \end{tabular}
        \end{sanseriffont}
        \fbox{
            \begin{sanseriffont}
                \begin{tabular}{@{}p{0.65\textwidth}@{}}
                    \textbf{Reference B:} in comparison with the study of there is again \hlomma{enlargement of the cardiac silhouette} with extensive \underline{\hlommb{bilateral pleural effusions}} and \underline{\hlommc{compressive atelectasis}} combined with \underline{\hlommd{pulmonary vascular congestion}}. \\
            \addlinespace
            \textbf{\textsc{ICon} \textit{w/o} \textsc{Zoom}:} as compared to the previous radiograph there is no relevant change. \hlomma{moderate cardiomegaly} with \underline{\hlommb{bilateral pleural effusions}} and subsequent areas of atelectasis. moderate pulmonary edema. no newly appeared focal parenchymal opacity suggesting pneumonia. \\
            \addlinespace
            \textbf{\textsc{ICon}:} as compared to the previous radiograph there is unchanged evidence of \hlomma{moderate cardiomegaly} with \underline{\hlommd{pulmonary vascular congestion}} and moderate pulmonary edema. unchanged moderate \underline{\hlommb{bilateral pleural effusions}} with areas of \underline{\hlommc{compressive atelectasis}} at the lung bases. no new parenchymal opacities. no pneumothorax. \\
                \end{tabular}
            \end{sanseriffont}
        }
    } 
	\caption{A case study of \textsc{ICon} on two semantically equivalent cases (i.e., Case A and Case B), given their radiographs and lesions. Spans with the same color (\hlomma{\textit{Cardiomegaly}}, \hlommb{\textit{Pleural Effusion}}, \hlommc{\textit{Atelectasis}}, and \hlommd{\textit{Edema}}) represent the same positive observation. Consistent and accurate outputs are highlighted with \underline{underline}.}
    \label{figure: case_study}
\end{figure*}
\begin{figure}[hpbt]
	\centering
    \setlength\belowcaptionskip{\fmargin}
    \resizebox{\columnwidth}{!}{
        \begin{sanseriffont}
            \begin{tabular}{@{}p{0.38\columnwidth}@{}}
            \centerline{ \includegraphics[width=0.9\linewidth]{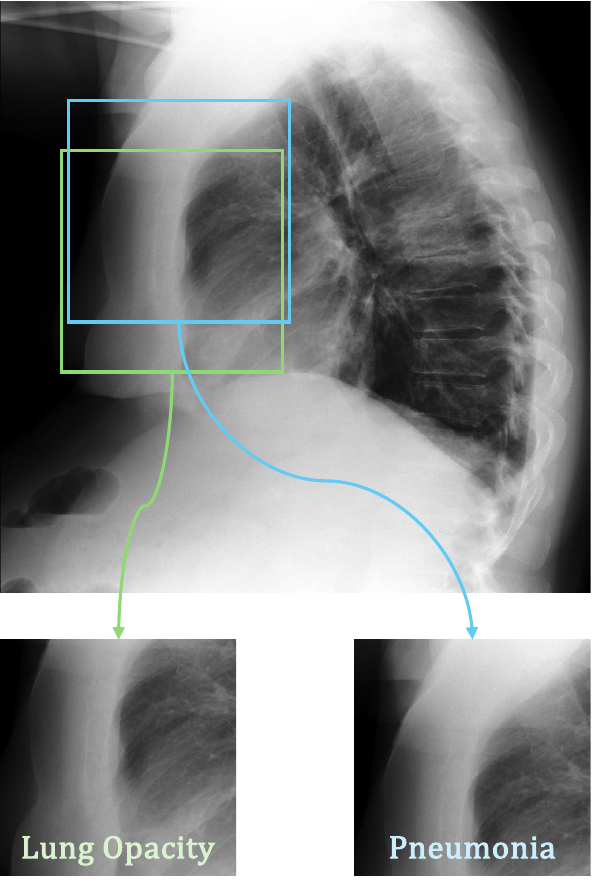}} \\
                \textcircled{1} \hlomme{Pneumonia/False Neg} \\
                \textcircled{2} \hlommf{Lung Opacity/False POS} \\
            \end{tabular}
        \end{sanseriffont}
        \fbox{
            \begin{sanseriffont}
                \begin{tabular}{@{}p{0.62\columnwidth}@{}}
                    \textbf{Reference:} pa and lateral views of the chest. there are new opacities in the superior segment of the left lower lobe and in the right lower lobe most consistent with \textcircled{1} \hlomme{multifocal pneumonia}. no pleural effusion or pneumothorax. cardiomediastinal and hilar contours are normal. \\
                    \addlinespace
                    \textbf{\textsc{ICon}:} \dots\dots the heart size remains unchanged and is within normal limits. \dots the pulmonary vasculature is not congested. no signs of acute or chronic parenchymal infiltrates are present and the lateral and posterior pleural sinuses are free. no pneumothorax in the apical area on frontal view. when comparison is made with the next preceding study there is a new area of \textcircled{2} \hlommf{increased opacity} in the left.
                \end{tabular}
            \end{sanseriffont}
        }
    }
	\caption{An error case produced by \textsc{ICon}. The \hlomme{span} and the \hlommf{span} denote false negative and false positive observations, respectively.}
    \label{figure: error_analysis}
\end{figure}
\noindent\textbf{Ablation Results.} The ablation results for \textsc{MIMIC-ABN} and \textsc{MIMIC-CXR} are listed in Table \ref{table: consistency_results} and Table \ref{table: mimic_ablation}. We study three variants: (1) \textit{w/o} \textsc{Zoom}, where all components are removed, (2) \textit{w/o} \textsc{Inspect}, where both the \textsc{Inspector} and \textsc{Mixup} are removed, and (3) \textit{w/o} \textsc{Mixup}, where only \textsc{Mixup} is removed. The performance of the variant \textit{w/o} \textsc{Zoom} drops significantly for both datasets, while the variant \textit{w/o} \textsc{Inspect} achieves competitive results in terms of CE scores. This suggests that the \textsc{Zoomer} effectively extracts lesions and provides relevant abnormal information for report generation. In addition, the variant \textit{w/o} \textsc{Mixup} further improves performance, demonstrating the effectiveness of the \textsc{Inspector} in transforming concise lesion information into precise reports. Moreover, introducing lesion-aware mixup augmentation strengthens the consistency of generated outputs, indicating the overall effectiveness of \textsc{ICon}.

\subsection{Qualitative Analysis}
\textbf{Case Study.} Figure \ref{figure: case_study} showcases two semantically equivalent cases, i.e., Case A and Case B, extracted from the test set of \textsc{MIMIC-CXR}. In both instances, \textsc{ICon} successfully identifies abnormal observations (e.g., \textit{Cardiomegaly}, \textit{Pleural Effusion}, and \textit{Atelectasis}) and generates consistent phrases including "\textit{pulmonary vascular congestion}", "\textit{bilateral pleural effusions}", and "\textit{compressive atelectasis}." Conversely, the variant \textit{w/o} \textsc{Zoom} fails to produce these descriptions in Case A. This demonstrates that \textsc{Zoomer} plays a crucial role in identifying lesions and highlights the ability of the mixup augmentation to ensure the alignment of lesions with their corresponding attributes.

\noindent\textbf{Error Analysis.} Figure \ref{figure: error_analysis} presents an error case produced by \textsc{ICon}. Although \textsc{Zoomer} successfully identifies \textit{Pneumonia} in the given radiographs, the \textsc{Generator} fails to realize it into descriptions like "\textit{multifocal pneumonia}" (i.e., a false negative observation). We notice that the region of this observation is inaccurately identified. Additionally, \textsc{Zoomer} outputs a false positive observation \textit{Lung Opacity}, leading to an inaccurate phrase "\textit{increased opacity}". To mitigate this issue, a better \textsc{Zoomer} trained with larger datasets could be beneficial.

\section{Related Works}
Radiology report generation \cite{coatt, hrgr, jing-etal-2019-show} has gained much attention. Prior research has either devised various memory mechanisms to record key information \cite{r2gen, r2gencmn, cmm-rl, metrans, zhao-etal-2023-normal} or proposed different learning methods to enhance performance \cite{ca, cmcl, ppked}. In addition, \citet{mia, dcl, kiut, yan-etal-2023-style} proposed utilizing knowledge graphs for report generation. \citet{clinical_reward, clinically_coherent, fact_ent, coplan, delbrouck-etal-2022-improving} designed various rewards for reinforcement learning to improve clinical accuracy. \citet{rgrg} proposed an explainable framework for report generation. \citet{organ} introduced observations to improve factual accuracy. \citet{kale-etal-2023-replace} proposed a template-based approach to improve the quality and accuracy of radiology reports. Additionally, \citet{ramesh2022improving, bannur2023learning, recap, serra-etal-2023-controllable} focused on exploring the temporal structure. \citet{wang-etal-2023-fine, wang-etal-2023-self-training} utilized CLIP \cite{clip} to bridge the modality gap. Mixup is also closely related to this research \cite{zhang2018mixup}, and this method has been extensively adopted in NLP research \cite{sun-etal-2020-mixup, yoon-etal-2021-ssmix, yang2022enhancing}. Although consistency has been studied in many domains \cite{10.1016/j.artint.2013.02.001, ribeiro-etal-2019-red, camburu2019make, elazar-etal-2021-measuring}, it remains unexplored in medical report generation.
\section{Conclusion and Future Works}
In this paper, we propose \textsc{ICon}, comprising three components to improve both accuracy and inter-report consistency. \textsc{ICon} first extracts lesions and then matches fine-grained attributes for report generation. A lesion-aware mixup method is devised for attribute alignment. Experimental results on three datasets demonstrate the effectiveness of \textsc{ICon}. In the future, we plan to explore incorporating large language models (LLMs) into our framework, given their advanced capabilities in planning and generation, to further enhance the performance of radiology report generation. Leveraging the strengths of LLMs could provide more refined signals to enhance the performance of \textsc{ICon}.

\section*{Limitations}
Although \textsc{ICon} can improve the consistency of radiology report generation, it still exhibits some limitations. Since our lesion extraction method is based on image labels, training such a model requires annotations for images. However, obtaining these annotations can be challenging in some medical settings. Recent advances in foundation vision models \cite{Kirillov_2023_ICCV} and open-set learning \cite{zara2023autolabel} could be a potential direction to address this issue. Additionally, image labels are coarse-grained, so the overall accuracy is likely to be lower than when using fine-grained labels (e.g., bounding boxes). Moreover, since our framework consists of two stages, prediction errors can propagate through the pipeline, making the final performance of our framework largely dependent on Stage 1. Reinforcement learning \cite{coplan} that takes factual improvement as a reward could be a solution to optimize the framework in an end-to-end manner.

\section*{Ethics Statement}
The \textsc{IU X-ray} \cite{iu_xray}, \textsc{MIMIC-ABN} \cite{mimic_abn}, and \textsc{MIMIC-CXR} \cite{mimic_cxr} datasets are publicly available and have been automatically de-identified to protect patient privacy. Our goal is to enhance the inter-report consistency of radiology report generation systems. Despite the substantial improvement of our framework over SOTA baselines, the performance still lags behind the requirements for real-world deployment and could lead to unexpected failures in untested environments. Thus, we urge readers of this paper and potential users of this system to cautiously check the generated outputs and seek expert advice when using it.

\section*{Acknowledgments}
This work was supported in part by the National Natural Science Foundation of China (82272086, 82102189, 82272086, 62076212) and the Research Grants Council of Hong Kong (15207920, 15207821, 15207122).
\bibliography{acl2024}
\appendix
\section{Appendix}
\subsection{Abnormal Observation Statistics}\label{appendix: obs_stat}
The abnormal observation statistics of \textsc{MIMIC-ABN}, \textsc{MIMIC-CXR}, and \textsc{IU X-ray} are listed in Table \ref{table: obs_stat}. 
\begin{table}[hpbt]
    \centering
    \resizebox{\linewidth}{!}{
    \begin{tabular}{l|l|l|l}
    \Xhline{2\arrayrulewidth}
    \textbf{\#Observation} & \textbf{\textsc{MIMIC-ABN}} & \textbf{\textsc{MIMIC-CXR}} & \textbf{\textsc{IU X-ray}} \\\hline
    \textit{No Finding} & 5002/32/22 & 64,677/514/229 &  744/108/318\\
    \textit{Cardiomegaly} & 16,312/118/244 & 70,561/514/1,602 & 244/38/61 \\
    \textit{Pleural Effusion} & 10,502/80/186 & 56,972/477/1,379 & 60/13/15 \\
    \textit{Pneumothorax} & 1,452/24/4 & 8,707/62/106 & 9/2/5 \\
    \textit{Enlarged Card.} & 5,202/40/90 & 49,806/413/1,140 & 159/29/28 \\
    \textit{Consolidation} & 4,104/36/96 & 14,449/119/384 & 17/1/3 \\
    \textit{Lung Opacity} & 22,598/166/356 & 67,714/497/1,448 & 295/35/57 \\
    \textit{Fracture} & 4,458/32/76 & 11,070/59/232 & 84/6/15 \\
    \textit{Lung Lesion} & 5,612/54/112 & 11,717/123/300 & 85/14/17 \\
    \textit{Edema} & 8,704/76/168 & 33,034/257/899 & 28/2/7 \\
    \textit{Atelectasis} & 19,132/134/220 & 68,273/515/1,210 & 143/15/37 \\
    \textit{Support Devices} & 9,886/58/196 & 60,455/450/1,358 & 89/20/16  \\
    \textit{Pneumonia} & 17,826/138/260 & 23,945/184/503 & 20/2/1  \\
    \textit{Pleural Other} & 2,850/30/62 & 7,296/70/184 & 32/4/7\\
    \Xhline{2\arrayrulewidth}
    \end{tabular}}
    \caption{Observation distribution in train/valid/test split of three datasets. \textit{Enlarged Card.} refers to \textit{Enlarged Cardiomediastinum}.}
    \label{table: obs_stat}
\end{table}

\subsection{Attributes of Observations}\label{appendix: attribute}
We list top-5 attributes for each observation for a better understanding in Table \ref{table: obs-attribute}.
\begin{table}[hpbt]
    \centering
    \resizebox{\linewidth}{!}{
    \begin{tabular}{l|l}
    \Xhline{2\arrayrulewidth}
    \textbf{Observation} & \textbf{Top-5 Attributes} \\\hline
    \textit{Cardiomegaly} & cardiomegaly, borderline, moderately, severely, mildly \\
    \textit{Pleural Effusion} & layering, subpulmonic, thoracentesis, trace, small\\
    \textit{Pneumothorax} & hydropneumothorax, apical, tiny, tension, component \\
    \textit{Enlarged Card.} & mediastinum, widening, contour, widened, lymphadenopathy \\
    \textit{Consolidation} & consolidative, collapse, underlying, developing, consolidations \\
    \textit{Lung Opacity} & opacification, opacifications, patchy, heterogeneous, scarring \\
    \textit{Fracture} & healed, fractured, healing, nondisplaced, posterolateral \\
    \textit{Lung Lesion} & nodular, nodule, mass, nodules, mm \\
    \textit{Edema} & indistinctness, asymmetrical, haziness, asymmetric, interstitial \\
    \textit{Atelectasis} & atelectatic, atelectasis, collapsed, subsegmental, collapse \\
    \textit{Support Devices} & sidehole, carina, coiled, tunneled, duodenum \\
    \textit{Pneumonia} & infectious, infection, atypical, supervening, developing \\
    \textit{Pleural Other} & fibrosis, thickening, biapical, blunting, scarring \\
    \Xhline{2\arrayrulewidth}
    \end{tabular}}
    \caption{Top-5 attributes for each observation.}
    \label{table: obs-attribute}
\end{table}

\subsection{Other Preprocessing Details}\label{appendix: data_preprocessing}
We adopt the same preprocessing setup used in \citet{r2gen}, and the minimum count of each token is set to 3/3/10 for \textsc{IU X-ray}/\textsc{MIMIC-ABN}/\textsc{MIMIC-CXR}, respectively. Other tokens are replaced with a special token \texttt{<unk>}.

\subsection{Additional Implementation Details}\label{appendix: s1_impl}
For Stage 1, all three datasets use the same hyper-parameters for training \textsc{Zoomer}, with a learning rate of $1e-4$, batch size of $128$, and dropout rate of $0.1$, and the number of training epochs is adjusted accordingly. We train \textsc{Zoomer} for 5, 10, and 15 epochs on \textsc{MIMIC-CXR}, \textsc{MIMIC-ABN}, and \textsc{IU X-ray}, respectively. During training, several data augmentation methods are applied. The input resolution of Swin Transformer is $256\times 256$, and we first resize an image to $288\times 288$, and then randomly crop it to $256\times 256$ with random horizontal flip. All experiments are conducted using one NVIDIA-3090 GTX GPU. For Stage 2, no data augmentation is applied, and we conduct experiments on \textsc{MIMIC-ABN} and \textsc{IU X-ray} using two NVIDIA-3090 GTX GPUs, and on \textsc{MIMIC-CXR} using four NVIDIA-V100 GPUs, both with half precision. Our model has 328.38M trainable parameters, and the implementations are based on the HuggingFace’s Transformers \cite{huggingface}. Here are the pretrained models we used:
\begin{itemize}[noitemsep,topsep=2pt]
    \item Small version of Swin Transformer V2: \url{https://huggingface.co/microsoft/swinv2-small-patch4-window8-256}
    \item Tiny version of Swin Transformer V2: \url{https://huggingface.co/microsoft/swinv2-tiny-patch4-window8-256}
    \item Base Version of Biomedical BART: \url{https://huggingface.co/GanjinZero/biobart-v2-base}
\end{itemize}

\begin{table}[t]
\centering
    \resizebox{\linewidth}{!}
    {
    \begin{tabular}{l|ccc|ccc}
    \Xhline{2\arrayrulewidth}
    \multirow{2}{*}{\textbf{Observation}} & \multicolumn{3}{c|}{{\textbf{Image Classification}}} &\multicolumn{3}{c}{{\textbf{Report Classification}}} \\\cline{2-7}
    & \textbf{P} & \textbf{R} & \textbf{F${_1}$} & \textbf{P} & \textbf{R} & \textbf{F${_1}$} \\\hline
    \textit{Enlarged Card.} & $0.426$ & $0.540$ & $0.476$ & $0.442$ & $0.525$ & $0.428$ \\
    \textit{Cardiomegaly} & $0.635$ & $0.838$ & $0.722$ & $0.630$ & $0.822$ & $0.714$ \\
    \textit{Lung Opacity} & $0.535$ & $0.725$ & $0.616$ & $0.542$ & $0.563$ & $0.552$\\
    \textit{Lung Lesion} & $0.318$ & $0.187$ & $0.235$ & $0.321$ & $0.177$ & $0.228$ \\
    \textit{Edema} & $0.471$ & $0.851$ & $0.607$ & $0.464$ & $0.784$ & $0.583$ \\
    \textit{Consolidation} & $0.283$ & $0.227$ & $0.251$ & $0.275$ & $0.162$ & $0.204$ \\
    \textit{Pneumonia} & $0.367$ & $0.396$ & $0.381$ & $0.341$ & $0.350$ & $0.345$ \\
    \textit{Atelectasis} & $0.541$ & $0.660$ & $0.595$ & $0.539$ & $0.620$ & $0.577$\\
    \textit{Pneumothorax} & $0.392$ & $0.481$ & $0.432$ & $0.400$ & $0.444$ & $0.421$ \\
    \textit{Pleural Effusion} & $0.719$ & $0.842$ & $0.776$ & $0.721$ & $0.827$ & $0.770$\\
    \textit{Pleural Other} & $0.289$ & $0.440$ & $0.349$ & $0.295$ & $0.315$ & $0.304$ \\
    \textit{Fracture} & $0.266$ & $0.198$ & $0.227$ & $0.225$ & $0.164$ & $0.190$\\
    \textit{Support Devices} & $0.747$ & $0.850$ & $0.795$ & $0.785$ & $0.784$ & $0.785$\\
    \textit{No Finding} & $0.366$ & $0.459$ & $0.407$ & $0.263$ & $0.535$ & $0.352$\\\hline
    Macro Average & $0.454$ & $0.550$ & $0.491$ & $0.445$ & $0.505$ & $0.464$ \\
    \Xhline{2\arrayrulewidth}
    \end{tabular}
    }
    \caption{Experimental results of each observation on the \textsc{MIMIC-CXR} dataset.}
    \label{table: s2_obs}
\end{table}

\begin{table}
\centering
    \resizebox{\linewidth}{!}
    {
    \begin{tabular}{l|ccc|ccc}
    \Xhline{2\arrayrulewidth}
    \multirow{2}{*}{\textbf{Model}} & \multicolumn{3}{c|}{{\textbf{MIMIC-ABN}}} & \multicolumn{3}{c}{{\textbf{MIMIC-CXR}}}  \\\cline{2-7}
    & \textbf{P} & \textbf{R} & \textbf{F${_1}$} & \textbf{P} & \textbf{R} & \textbf{F${_1}$} \\\hline
    \textsc{R2Gen} & $0.340$ & $0.413$ & $0.348$ & $0.390$ & $0.336$ & $0.337$  \\
    \textsc{R2GenCMN} & $0.360$ & $0.363$ & $0.336$ & $0.358$ & $0.276$ & $0.290$  \\
    \textsc{RGRG} & $-$ & $-$ & $-$  & $0.461$ & $0.475$ & $0.447$ \\
    \textsc{ORGan} & $0.418$ & $0.471$ & $0.412$ & $0.493$ & $0.560$ & $0.493$  \\
    \textsc{Recap} & $0.366$ & $0.468$ & $0.382$  & $0.447$ & $0.558$ & $0.464$ \\\hline
    \textsc{ICon} & $0.512$ & $0.428$ & $0.436$ & $0.513$ & $0.597$ & $0.522$ \\
    \textsc{ICon} \textit{w/o} \textsc{Zoom} & $0.397$ & $0.406$ & $0.372$ & $0.440$ & $0.362$ & $0.373$  \\
    \textsc{ICon} \textit{w/o} \textsc{Inspect} & $0.430$ & $0.479$ & $0.424$ & $0.506$ & $0.553$ & $0.500$  \\
    \textsc{ICon} \textit{w/o} \textsc{Mix-up} & $0.433$ & $0.509$ & $0.438$ & $0.507$ & $0.590$ & $0.517$  \\
    \Xhline{2\arrayrulewidth}
    \end{tabular}
    }
    \caption{Example-based CE results on the MIMIC-ABN and MIMIC-CXR datasets.}
    \label{table: example_based_ce}
\end{table}

\subsection{Lesion Extraction}\label{appendix: extract_region}
There are two steps in extraction lesions: candidate generation and candidate classification. Given an image with a resolution of 1024$\times$1024, padding if needed, we apply a sliding window of 384$\times$384, with a step size of 128 to extract candidates for classification. This operation results in $36$ regions. Then, each region is fed into the \textsc{Zoomer} for classification, and only the top-$1$ region is selected for each observation. Note that before extracting lesions, each input case is first assigned with their observations by \textsc{Zoomer}, and as a result, the number of lesions corresponds to the number of observations.

The \textit{No Finding} observation is excluded for lesion extraction, as it estimates the overall conditions of a patient, which makes it difficult to locate at specific regions.

\subsection{Justifications for Additional Data Processing}\label{appendix:justification}
\textbf{Justification for Using Historical Records.} As stated in \citet{bannur2023learning,recap}, without historical information, it is unreasonable to generate reports with comparisons between two consecutive visits and will lead to hallucinations \cite{ramesh2022improving}. As a result, we include historical records as context information for report generation.

\noindent\textbf{Justification for Using All Views.} Prior research \cite{r2gen,r2gencmn, organ, recap} treated different views of radiographs in one visit as different samples. However, this is unreasonable to generate a report with only one view position, since different diseases could be observed from different view positions. For example, most of the devices can not be observed from a Lateral view. Given a lateral view radiograph, writing a sentence of "\textit{A right chest tube is in unchanged position.}" is unreasonable.

In addition, some reports describe how many views are provided at the beginning, e.g., "\textit{PA and lateral views are provided.}" Above all, we have justified reasons to use all the views in one visit of a patient to generate the target report. Note that previous work treated each image as a sample and their settings have more samples than ours. For a fair comparison, each generated output of a study with $L$ images is duplicated $L$ times so that the number of samples in evaluation is consistent with previous research.

\begin{table}[t]
\centering
    {
    \begin{tabular}{l|ccc}
    \Xhline{2\arrayrulewidth}
    \textbf{Dataset} & \textbf{P} & \textbf{R} & \textbf{F${_1}$} \\\hline
    \textsc{IU X-ray} & $0.223$ & $0.243$ & $0.225$ \\
    \textsc{MIMIC-ABN} & $0.379$ & $0.472$ & $0.411$ \\
    \textsc{MIMIC-CXR} & $0.454$ & $0.550$ & $0.491$ \\
    \Xhline{2\arrayrulewidth}
    \end{tabular}
    }
    \caption{Abnormal observation prediction results of \textsc{Zoomer} at Stage 1. Results on the \textsc{IU X-ray} dataset are only provided for reference.}
    \label{table: s1_zoomer}
\end{table}
\label{sec:appendix}

\end{document}